\documentclass[10pt,twocolumn,letterpaper]{article}

\usepackage{cvpr}
\usepackage{times}
\usepackage{epsfig}
\usepackage{graphicx}
\usepackage{amsmath}
\usepackage{amssymb}

\usepackage{bbm}
\usepackage{tabularx}
\usepackage{lscape}
\usepackage{rotating}
\usepackage{adjustbox}

\usepackage[pagebackref=true,breaklinks=true,letterpaper=true,colorlinks,bookmarks=false]{hyperref}

\cvprfinalcopy % *** Uncomment this line for the final submission

\def\cvprPaperID{****} % *** Enter the CVPR Paper ID here

\ifcvprfinal\fi
\begin{document}

%%%%%%%%% TITLE
\title{NeRF applied to satellite imagery for surface reconstruction}

\author{Federico Semeraro\thanks{Random order. All authors made equal contribution.}, \, Yi Zhang$^*$, \, Wenying Wu$^*$, \, Patrick Carroll$^*$ \\
Georgia Institute of Technology\\
{\tt\small \{fsemeraro6, yzhang3416, wwu393, pcarroll7\}@gatech.edu}}

\maketitle

%------------------------------------------------------------------------
\begin{abstract}

We present Surf-NeRF, a modified implementation of the recently introduced Shadow Neural Radiance Field (S-NeRF) model. This method is able to synthesize novel views from a sparse set of satellite images of a scene, while accounting for the variation in lighting present in the pictures. The trained model can also be used to accurately estimate the surface elevation of the scene, which is often a desirable quantity for satellite observation applications. S-NeRF improves on the standard Neural Radiance Field (NeRF) method by considering the radiance as a function of the albedo and the irradiance. Both these quantities are output by fully connected neural network branches of the model, and the latter is considered as a function of the direct light from the sun and the diffuse color from the sky. The implementations were run on a dataset of satellite images, augmented using a zoom-and-crop technique. A hyperparameter study for NeRF was carried out, leading to intriguing observations on the model's convergence. Finally, both NeRF and S-NeRF were run until 100k epochs in order to fully fit the data and produce their best possible predictions. The code related to this article can be found at \href{https://github.com/fsemerar/surfnerf}{https://github.com/fsemerar/surfnerf}.

\end{abstract}

%------------------------------------------------------------------------
\section{Introduction}

A revolution is currently underway in the aerospace field driven by the miniaturization of satellites, combined with a significant decrease in the cost to access space, thanks to the competition between launching providers. These trends have triggered the emergence of a myriad of new private companies looking to exploit this new frontier by affordably sending CubeSats into orbit. One of the most commonly pursued applications is Earth observation, with the purpose of gaining insights from satellite imaging. One of the immediate goals of remote sensing is the reconstruction of the surface, from which several properties can be calculated (change detection, land classification, etc.). Even though direct surface sensing technologies exist, such as Light Detection And Ranging (LiDAR) or radar satellite imaging, they are often more expensive to manufacture, involved to maintain, and require higher energy than a simple passive camera. For this reason, photogrammetry techniques to extract information from normal high resolution images are currently in high demand.

\begin{figure}[!htb]
      \centerline{\includegraphics[width=0.8\linewidth]{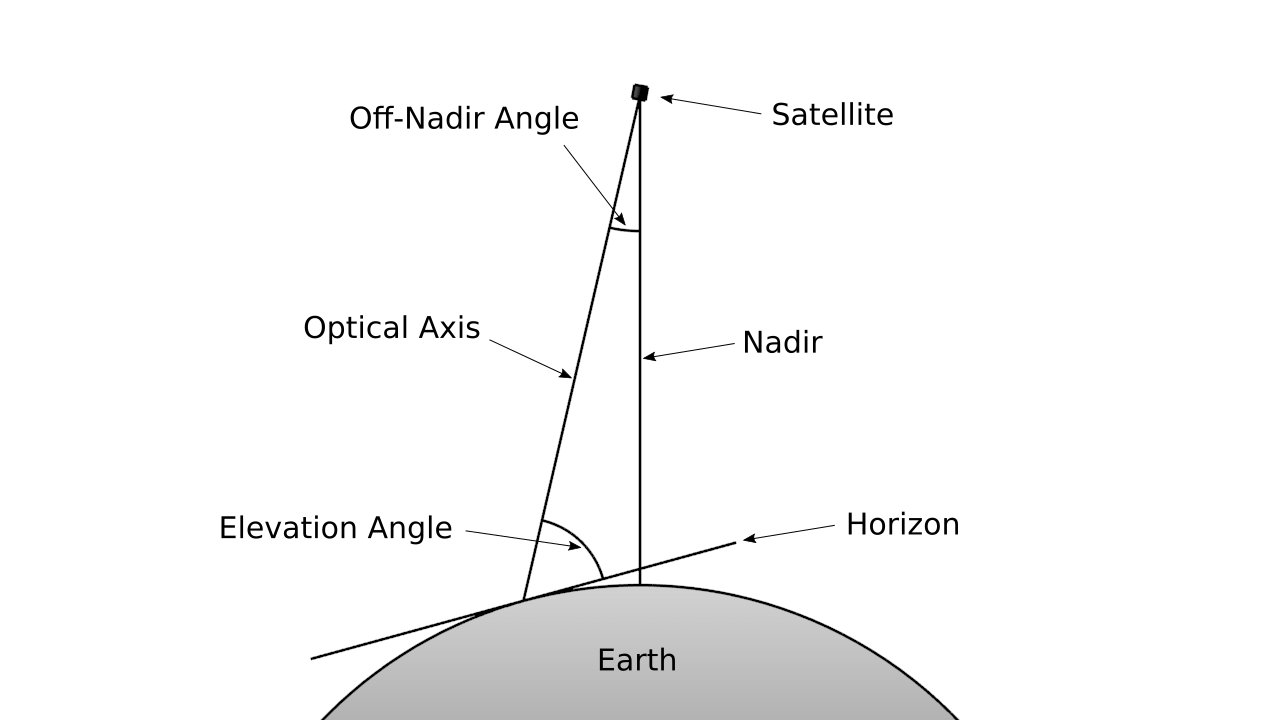}}
  \caption{Important angles for satellite imaging \label{fig:sat_angles}}
\end{figure}

This is what motivated our study, which aims at employing state-of-the-art models in order to reconstruct the surface of the Earth from a sparse set of satellite images. The unique challenges of our application can be summarized as follows: first, differently from other NeRF applications, we only have images of a scene from above, with an \emph{off-nadir} viewing angle, shown in Figure \ref{fig:sat_angles}, of below $35^{\circ}$. In addition, we only have access to a limited number of images, 10-20 per scene, which were taken at different times throughout the year. This translates in strong non-correlations within each scene dataset due to different lighting conditions, as well as small scene variations (e.g. cars, vegetation, etc.).

\begin{figure}[!htb]
      \centerline{\includegraphics[width=\linewidth]{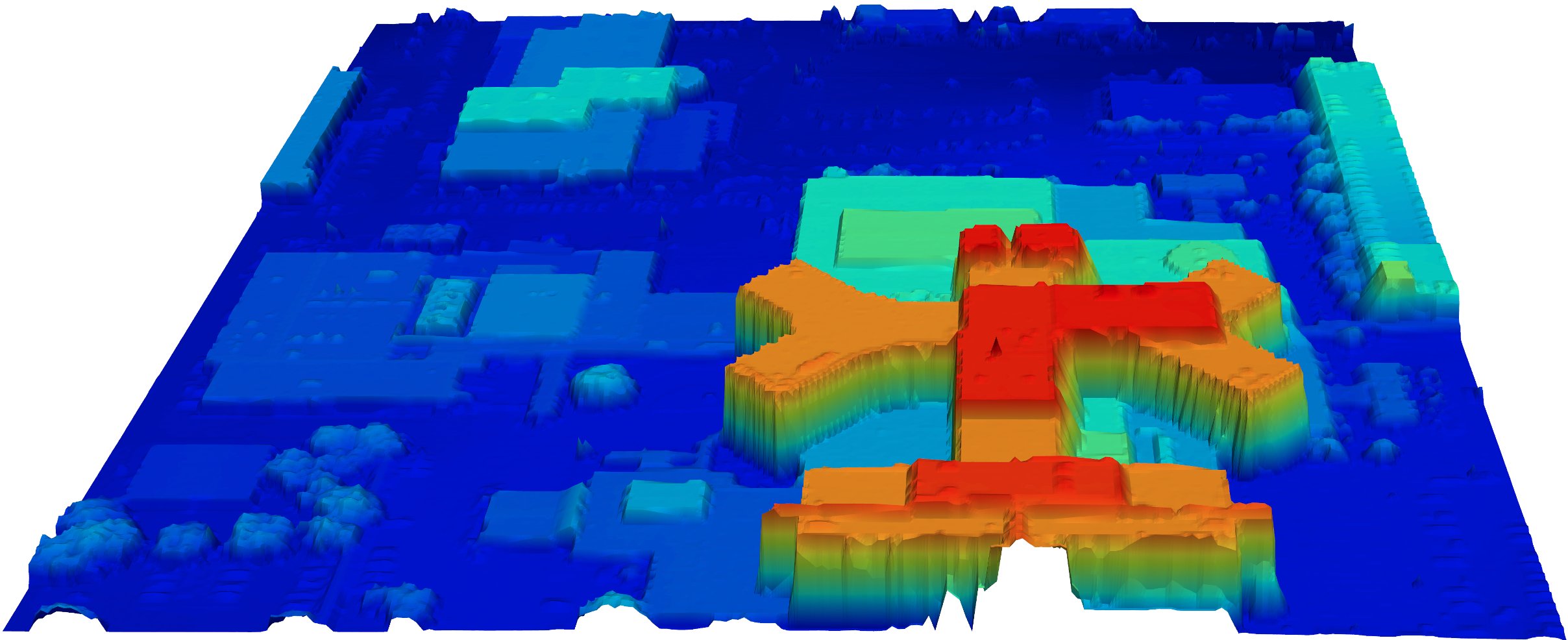}}
  \caption{Ground truth Digital Surface Model (DSM) for scene 068 of the Jacksonville dataset \label{fig:dsm}}
\end{figure}

In this project, we rework and implement a recently proposed technique called Shadow-NeRF \cite{derksen2021shadow}, based on a 3D computer graphics technique called Neural Radiance Fields (NeRF) \cite{mildenhall2020nerf}. We used the PyTorch implementation of NeRF \cite{lin2020nerfpytorch} as a starting point. The main difference introduced by S-NeRF is the direct modeling of the lighting condition, which is essential for its application to satellite imagery of urban scenes. For our investigation, we utilize the open-source image dataset of Jacksonville, provided by \cite{c6tm-vw12-19}, that was captured by DigitalGlobe's WorldView-3 satellite, orbiting at 617km altitude and taking images with pixel resolution of 0.3m. However, the original full resolution images are downsampled to a resolution of 0.6m in order to compare them with a ground truth LiDAR elevation map having resolution of 0.5m, which can be observed in Figure \ref{fig:dsm}.

%------------------------------------------------------------------------
\section{Related Work}

The construction of novel views and estimation of a depth map from a sparse set of images are long-standing problems in computer vision, that have been addressed in multiple ways. Several techniques, e.g. bundle adjustment \cite{triggs1999bundle} and Structure-from-Motion and \cite{andrew2001multiple}, have been devised to construct a point cloud representation of a scene. For the specific application to satellite imagery, more traditional methods rely on direct stereo matching of image pixels via the optimization of an energy function that encourages regular 3D models \cite{boykov2001fast}. 

Recently, the NeRF approach has taken the stage, thanks to the state-of-the-art results that it produces when synthesizing novel views. This method takes an alternative route than previous approaches by modeling the radiance field and the density of a scene as a continuous function using the weights of a Multi-Layer Perceptron (MLP). The real power of NeRF, as compared to previous deep learning attempts, is that the MLP training happens in a completely self-supervised manner, by using ray casting and traditional volume rendering to compare the predicted pixel color to the ground truth training images. Since its conception, several improvements have been proposed to cope with some of the shortcomings of the original method, making the NeRF approach usable for several real world applications, such as robotics and trajectory planning \cite{adamkiewicz2021vision, li20213d}.

One of the requirements to train the MLP is the knowledge of the camera pose, a.k.a the viewing angles $(\theta,\phi)$, at which the training images were taken. This requirement was relaxed by recent improvements proposed by several authors \cite{lin2021barf, wang2021nerf}. NeRF--,  for instance, is one of the improvements that tackles novel view synthesis without known camera poses or intrinsics. However, since satellite images almost always come with the camera intrinsics and extrinsics, there are no particular benefits in using NeRF--, but only additional computational complexity. Another interesting improvement is the relaxation of the number of images required to represent the scene. This was proposed in the Pixel-NeRF model \cite{yu2021pixelnerf}, which claims that a much lower number of images are necessary to train. Our attempt to use their models on satellite images or to merge their approaches into the S-NeRF model did not result in satisfactory results, possibly due to the challenging viewing angles that our images have.

Finally, a more relevant property for our context is the increase in robustness proposed by the NeRF in the Wild (NeRF-W) model \cite{martin2021nerf}, that explicitly aims at modeling the occlusions and non-correlations present within the training images using a generative latent optimization. NeRF-W separates the static and transient components throughout the training set, which would be particularly useful for satellite images in order to filter out transient elements, such as shadows, cars, vegetation differences. However, we decided to pursue the inclusion of the S-NeRF approach to the original NeRF instead, because it is more tailored towards satellite photography. By explicitly modeling the lighting condition, S-NeRF is able to train and bias its weights relying on the sun direction angle as an extra input to correct for the fact that satellite imagery data is often taken at various times of the day, causing different shadows to be cast on the ground, problematic for traditional NeRF approaches.

%-------------------------------------------------------------------------
\section{Approach}

% NeRF theory
\subsection{Neural Radiance Fields}

\begin{figure*}
  \centerline{\includegraphics[width=0.7\textwidth]{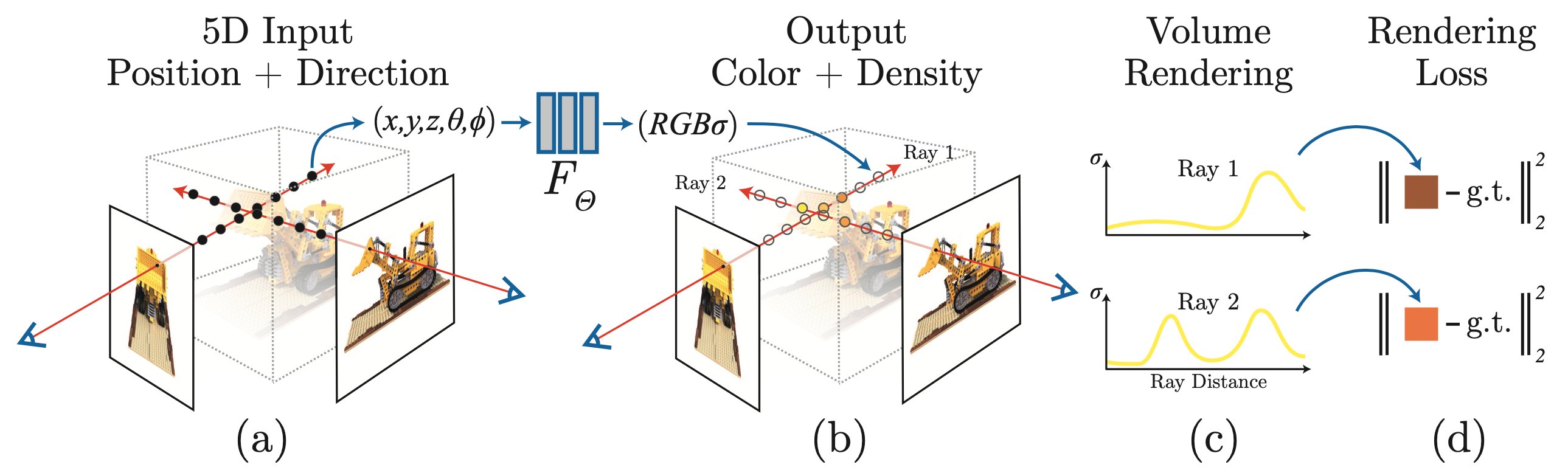}}
  \caption{Visual steps for the NeRF model \cite{mildenhall2020nerf}: (a) describes the ray sampling across the scene, which creates the 5D input for the MLP; (b) showcases the MLP outputs in the form of an estimated RGB color, which depends on both 3D location and viewing angle, and scene density $\sigma$; (c) outlines the differentiable volume rendering approach to alpha composite along each ray; finally (d) shows the self-supervised loss function, which compares the predicted color to the ground truth color of the training images \label{fig:nerf}}
\end{figure*}

The NeRF method attempts to model a scene as a continuous 5D radiance function approximated by the weights of a fully connected neural network, which we represent with the symbol $F_{\Theta}$. The inputs to the functions are the 3D locations $\boldsymbol{x} = (x,y,z)$ throughout the scene and the viewing angles $\boldsymbol{d} = (\theta, \phi)$, which are actually represented as unit vectors with three Cartesian components. Its outputs are a location-dependent volume density $\sigma(\boldsymbol{x})$ and a location- and view-dependent color $\boldsymbol{c}(\boldsymbol{x},\boldsymbol{d}) = (r,g,b)$.

As in the original NeRF paper, the MLP architecture is divided into two steps: first, 8 fully-connected layers, having 256 channels per layer and using ReLU activation functions, process the 3D input coordinates $\boldsymbol{x}$, and output $\sigma$ together with a feature vector. Next, the latter feature vector is appended to the camera viewing angles and passed to a single fully-connected layer, with 128 channels and ReLU activation, which outputs the RGB color. 

A loss function is constructed based on a volume rendering approach. The general idea is to compute alpha compositing along each ray and compare the predicted color against the ground truth from the corresponding training image. Alpha compositing can be described as: 

\begin{equation} \label{eq:alpha_compositing}
I_{ray} = \sum_{i=1}^{N} T_i \alpha_i \boldsymbol{c}_i
\end{equation}

where the predicted color $I$ is computed as the weighted sum of the colors $\boldsymbol{c}_i$ along the ray. The weights are given by the multiplication of the sampled transparency $T_i$ and the local opacity value $\alpha_i$. These two quantities are computed using the following equations: 

\begin{equation} \label{eq:alpha}
\alpha_i = 1 - e^{-\sigma_i \delta x_i}
\end{equation}

\begin{equation} \label{eq:transparency}
T_i = \prod_{j=0}^{i-1} (1-\alpha_j)
\end{equation}

Note that the opacity varies as $\alpha \in [0, 1]$, where an opacity of 0 would indicate a totally transparent material, whereas $\alpha=1$ would describe a fully opaque material.  $\alpha_i$ determines how much light is contributed by ray sample $i$. $T_i$ is modeled as the cumulative product of the transparency, i.e. the inverse opacity, from the camera to the current ray segment $i$.  The advantage of this formulation is that it is fully differentiable, enabling automatic back-propagation through the neural network.

At training time, as described in Figure \ref{fig:nerf}, a batch of pixels $N_b$ is selected at random from the training images. Ray marching is executed from these pixels by sampling in $N$ locations, each sample having coordinates $(x,y,z,\theta,\phi)$. These serve as input to the MLP, which then predicts the colors and density $(r,g,b,\sigma)$ of these sampled locations. Finally, the predicted color is computed along each ray according to the alpha compositing equations described earlier. This color prediction $\boldsymbol{c}_{ray}$ is then compared against the ground truth color $\boldsymbol{c}_{gt}$ from the training image using a simple total squared error loss:

\begin{equation} \label{eq:loss}
\mathcal{L} = \sum_{ray=1}^{N_b} \| \boldsymbol{c}_{ray} - \boldsymbol{c}_{gt} \|^2_2
\end{equation}

This loss is back-propagated through the MLP's weights in order to find a better fit for the training images. The NeRF method is meant to overfit the weights of a neural network to encode the scene itself. If a new scene needs to be rendered, an entirely new training run is required (one model per scene).

% S-NeRF theory
\subsection{S-NeRF}

\begin{figure}[!htb]
      \centerline{\includegraphics[width=0.95\linewidth]{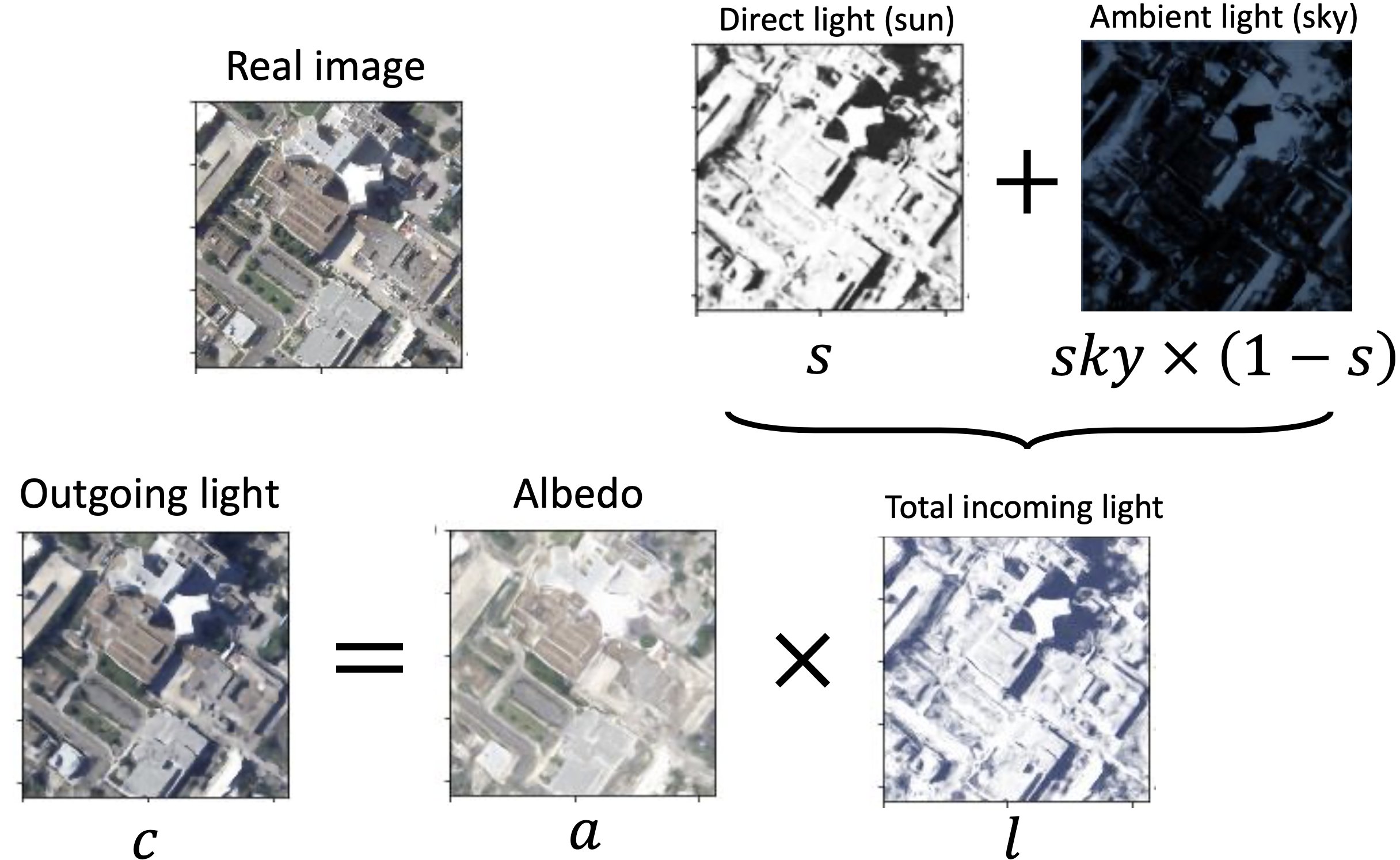}}
  \caption{S-NeRF irradiance model \cite{derksen2021shadow}: the outgoing light is modeled as the scene's albedo $\boldsymbol{a}$ times the total incoming light $l$. The latter quantity is modeled as the combination of the directional light source from the sun $s$ and the diffuse source from the sky \label{fig:snerf}}
\end{figure}

While NeRF is an emissive-only light transport model, S-NeRF considers the irradiance of a scene by approximating the outgoing light as the combination of the scene's albedo and the total incoming light, as shown in Figure \ref{fig:snerf}. This variation was introduced to overcome one of the main limitations of basic NeRF, namely the need of working with training images with constant lighting conditions. Since it is almost impossible to have satellite images with the same shadows, a more flexible model was required. 

S-NeRF introduces two new physical properties, which in practice are simply two new outputs of the MLP: $s(\boldsymbol{x}, \boldsymbol{\omega}_s) \in [0, 1]$ and $\boldsymbol{sky}(\boldsymbol{\omega}_s) \in [0, 1]^3$, both depending on a new input representing the sun direction $\boldsymbol{\omega}_s = (\theta_s, \phi_s)$. The first output, $s$, represents the amount of incoming solar directional light at each point in space, i.e. the visibility between the sampled points and the sun. The second output, $\boldsymbol{sky}$, is a color vector modelling the diffuse light source from the sky, which is considered the same throughout the scene. i.e. it only depends on the sun direction.

The lighting terms are considered as:

\begin{equation} \label{eq:radiance_snerf}
\boldsymbol{c}(\boldsymbol{x}, \boldsymbol{\omega}_s) = \boldsymbol{a}(\boldsymbol{x}) \cdot \ell (\boldsymbol{x}, \boldsymbol{\omega}_s)
\end{equation}

in which the irradiance $\ell$ is modelled as the weighted sum between $\boldsymbol{sky}$ and the white light source $\mathbbm{1}_3$:

\begin{equation} \label{eq:radiance_snerf}
\ell = s \mathbbm{1}_3 + (1-s) \boldsymbol{sky}
\end{equation}

Alpha compositing therefore becomes:

\begin{equation} \label{eq:alpha_compositing_snerf}
I_{ray} = \sum_{i=1}^{N} T_i \alpha_i \boldsymbol{a}_i \ell_i
\end{equation}

Note that this model does not account for specular reflections and takes the simplifying assumption that every shaded area is exposed to the same amount of light. 

Along the rays, the transparency $T_i$ is computed as described in Equation \ref{eq:transparency}. Two terms are then added to the original NeRF loss function: the first one penalizes the difference between $T_i$ and $s_i$, because along a solar ray the transparency should be equivalent to the solar light. The second term is $1 - L_1$ norm of $ws$, which holds the idea that all the solar light should be absorbed by the visible parts of the scene. These two parameters are modulated by $\lambda_s$, which describes the balance between shadow prediction and overall radiance accuracy. The S-NeRF loss function can be finally written as: 

\begin{multline} \label{eq:loss_snerf}
\mathcal{L} = \sum_{ray=1}^{N_b} \| \boldsymbol{c}_{ray} - \boldsymbol{c}_{gt} \|^2_2 + \\
\lambda_s \sum_{ray=1}^{N_b} \left( \sum_{i=1}^{N} (T_i - s_i)^2 + \left(1 - \sum_{i=1}^{N} w_i s_i \right) \right) 
\end{multline}

The network architecture that is used for our implementation of S-NeRF is a modified version of NeRF's MLP, with an initial branch of 8 layers and 256 channels that takes the positional information and outputs the density and the albedo $\boldsymbol{a} = (r,g,b)$. The 256 feature vector that is output by the first fully connected network branch is then concatenated with the solar direction $\boldsymbol{\omega}_s = (\theta_s, \phi_s)$ and the camera pose $\boldsymbol{d} = (\theta, \phi)$ and input into an extra 3 fully-connected layers, having 50 channels with ReLU activations between the layers and final Sigmoid activation. This branch outputs $s(\boldsymbol{x}, \boldsymbol{\omega}_s)$, the scalar sun visibility function. On the other hand, the solar direction alone is used as an input to another branch of the S-NeRF MLP with a single linear layer with 50 channels and Sigmoid activation. This network branch is used to output the sky color $\boldsymbol{sky} = (r,g,b)$. 

Our implementation of S-NeRF, however, employs positional encodings, while the original S-NeRF one uses SIREN activations \cite{sitzmann2020implicit} to preserve high frequency details. 

% positional encoding
\subsection{Positional encoding}

Deep neural networks have a natural bias towards preserving lower frequency information \cite{rahaman2019spectral}. This translates into blurry images without high frequency details. However, it has been shown that positional encodings can help an MLP to better fit the high frequencies in the training data. This is equivalent to mapping the inputs to a high dimensional space using periodic functions. Mathematically, this fixed mapping can be expressed as $F_{\Theta} = F'_{\Theta} \circ \gamma$, in which only $F'_{\Theta}$ is learned. The type of mapping that was proposed by the NeRF method is: 

\begin{equation} \label{eq:posenc}
\resizebox{.9\hsize}{!}{$\gamma(p) = (\sin{(2^0 \pi p)}, \cos{(2^0 \pi p)}, ... , \sin{(2^{L-1} \pi p)}, \cos{(2^{L-1} \pi p)})$}
\end{equation}

This encoding function is applied to each ray sample coordinate $(x,y,z,\theta,\phi)$. For the first three spatial coordinates, NeRF recommends using $L=10$, while for the viewing directions $L=4$. 

% ray sampling
\subsection{Ray sampling}

Along each ray, we sample $N$ times in order to evaluate the density and color at each location. However, densely sampling each ray using a fine spacing would be intractable and inefficient since only the locations containing material contribute to the radiance. For this reason, a hierarchical sampling approach is taken, in which two networks are considered: one performing stratified sampling in a coarse manner, whose predictions then guide a second network that samples more finely in the locations that matter. Note that both ray samples are used conjointly in the loss function. A small variation introduced by S-NeRF, tailored to aerial footage applications, aims at sampling based on the altitude. Since we work with satellite images taken hundreds of kilometers away, we only want to sample in the regions close to the Earth's surface. For this reason, a minimum and maximum altitude based on the LiDAR surface measurements are considered while sampling. 

% data augmentation
\subsection{Data augmentation}

\begin{figure}[!htb]
      \centerline{\includegraphics[width=0.8\linewidth]{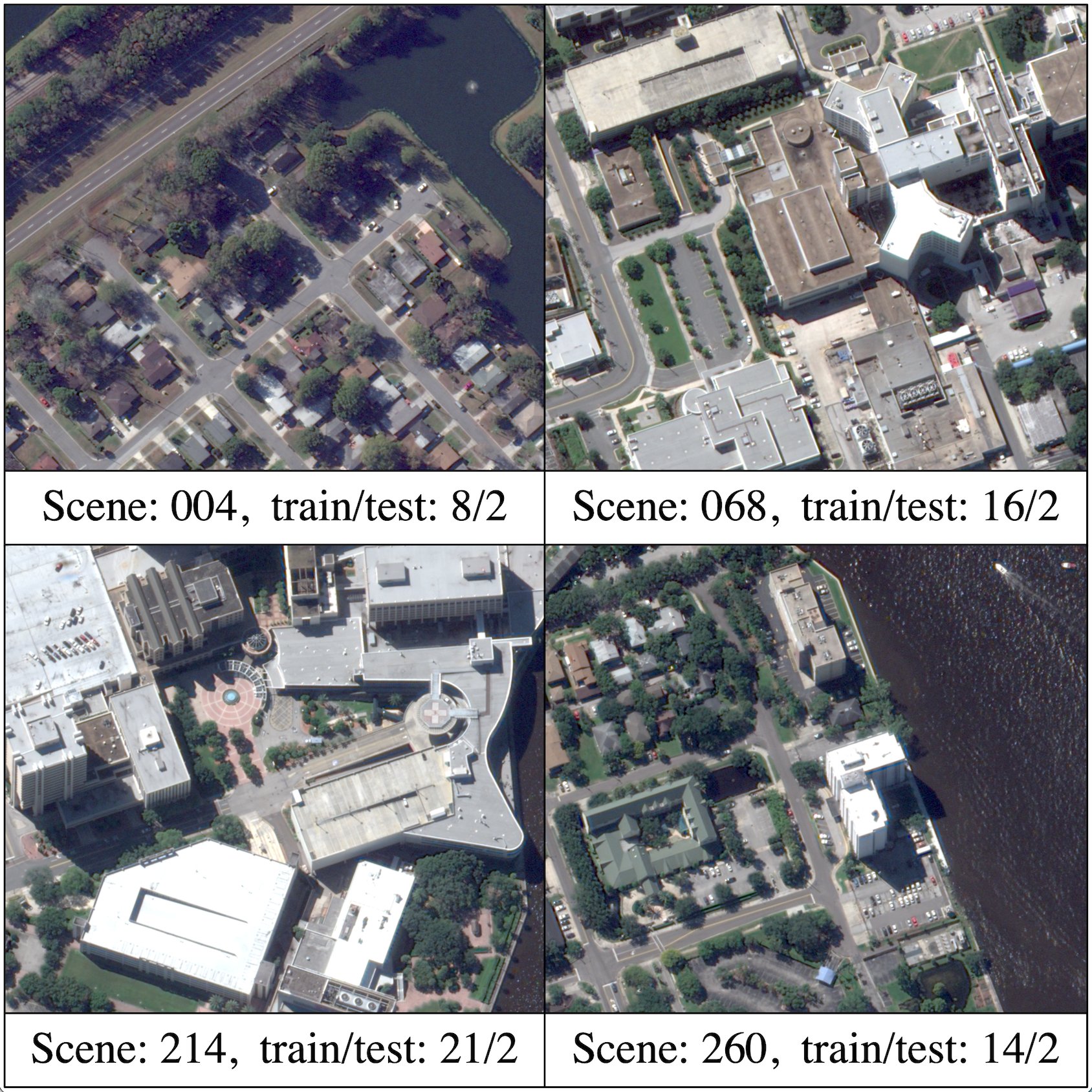}}
  \caption{Jacksonville scenes used in this study ($300$m$^2$ area) \label{fig:scenes}}
\end{figure}

In our project, we used the Jacksonville scenes from the original dataset, which was also used by the S-NeRF project. These satellite images were collected and open-sourced by the IEEE GRSS organization \cite{c6tm-vw12-19}. Four scenes were used for generating our results, each with a different number of images available, as it can be observed in Figure \ref{fig:scenes}. For each scene, 2 images were used for testing unseen viewing and lighting conditions.

Because of the limited number of images, we experimented with data augmentation techniques to generate more views for training purposes. However, due to the nature of our application, certain positional transformations do not work because they will likely alter the camera intrinsics. Given the fact that image magnification is directly proportional to the focal length of the camera, we decided to employ a zoom-and-crop approach to generate new images. Rasterio, a GDAL-based library, is able to deal with geo-spatial raster data, through which we were able to generate an additional set of GeoTIFF images with varying focal lengths. We also provided the option to apply Gaussian blurs to the augmented set.

%-------------------------------------------------------------------------
\section{Experimental Results and Discussion}

The evaluation of NeRF and S-NeRF was run on the four full-resolution scenes from the Jacksonville dataset, containing satellite images of $852\times852$ pixels. Model training was performed on NVIDIA V100 GPUs, taking around 100k iterations (roughly 20 hours) for the MLP to optimally fit the training images. The Adam optimizer was used to drive the weights to convergence, with a learning rate of $5e^{-4}$ decaying exponentially at a rate of $0.1$. Similarly to the original NeRF paper, we used a ray batch size of 4096, with each ray sampled $N_c=64$ times for the coarse network, and $N_c=128$ for the fine network. The metrics that were used to evaluate the model's performance are the Peak Signal-to-Noise Ratio (PSNR) and the Structural Similarity (SSIM) between the test images (i.e. held out views) and the model's predicted view. 

% Hyper-parameter tuning
\subsection{NeRF hyperparameter study}

\begin{figure}[!htb]
      \centerline{\includegraphics[width=\linewidth]{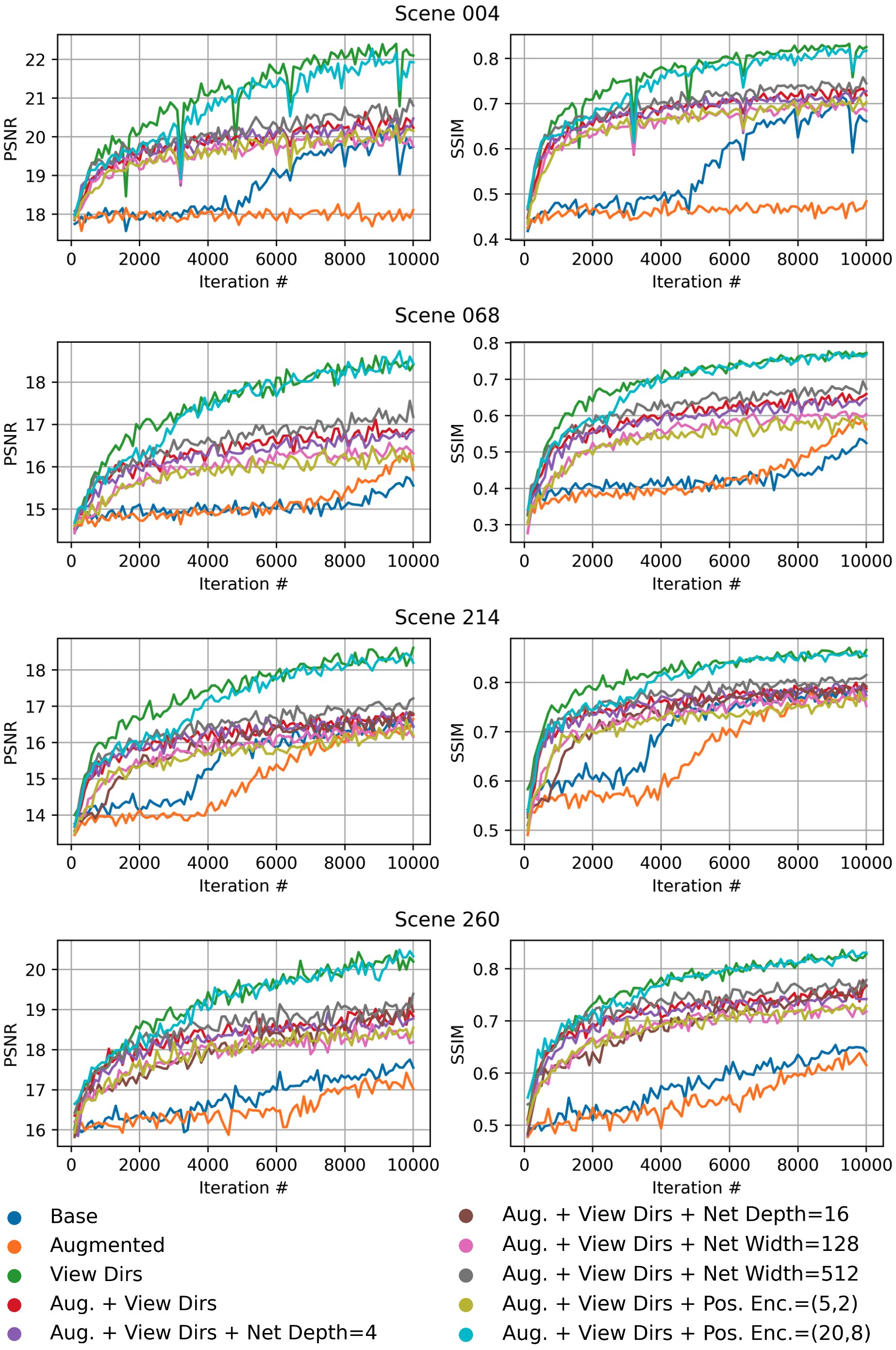}}
  \caption{Hyperparameter study for the basic NeRF model \label{fig:hyper}}
\end{figure}

Given the large amount of degrees of freedom available for tuning NeRF, only a subset of hyperparameters was investigated. This exploration is summarized in Figure \ref{fig:hyper}, which shows the convergence differences between the altered hyperparameters (colored lines), between the two metrics (left column for PSNR, right for SSIM), and the four scenes (rows). The number of epochs was truncated to 10,000 for practical purposes, since convergence trends can still be captured in the initial training regime. The qualitative results for these scenes can be found in Figure \ref{fig:qualitative_hyper}.

The most glaring aspect is that data augmentation did not help the performance, but instead hindered convergence in some cases. Under further scrutiny of the loss function, we realized that adding extra cropped sections to the training images do not provide any extra ground truth pixel colors, which are used to compare the predicted colors from the rays alpha compositing. The addition of redundant pixels may actually be the reason for the convergence slow-down. 

The second important observation is that varying the network depth and width did not have any significant effect on the convergence. On the other hand, the hyperparameter that seemed to add the strongest performance boost was the number of frequencies used in the positional encoding. This can be explained by the amount of very fine high frequency details present in satellite images of urban scenes, which the MLP struggles to learn. In addition, the inclusion of the viewing directions in the MLP inputs has a clearly beneficial effect on the model's convergence. 

Several unexpected aspects were noticed. First of all, the results that perform better in terms of the PSNR and SSIM metrics do not necessarily look better qualitatively. Some examples can be observed in Figure \ref{fig:batching_viewdirs} in the Appendix. This is especially true when deciding to use the viewing angles or not: including them in the MLP's inputs significantly improves the quantitative metrics, but the results look more noisy. This might be because of the extra degrees of freedom that the MLP has to deal with, i.e. including the viewing angles comes with the assumption that the colors may change depending on the prospective, which might push the MLP to vary them more strongly. Similarly, using a higher positional encoding also helps the metrics, but worsens the results qualitatively, possibly due to the same reason. Another important option that is left as a hyperparameter is whether to pick the batch of rays from all the training images at random, or whether to pick from one training image per iteration. This has an especially significant effect on the early model's convergence, as it can be noticed in Figure \ref{fig:batching_viewdirs} in the Appendix.

\subsection{Models performance and altitude estimation}

\begin{figure}[!htb]
      \centerline{\includegraphics[width=\linewidth]{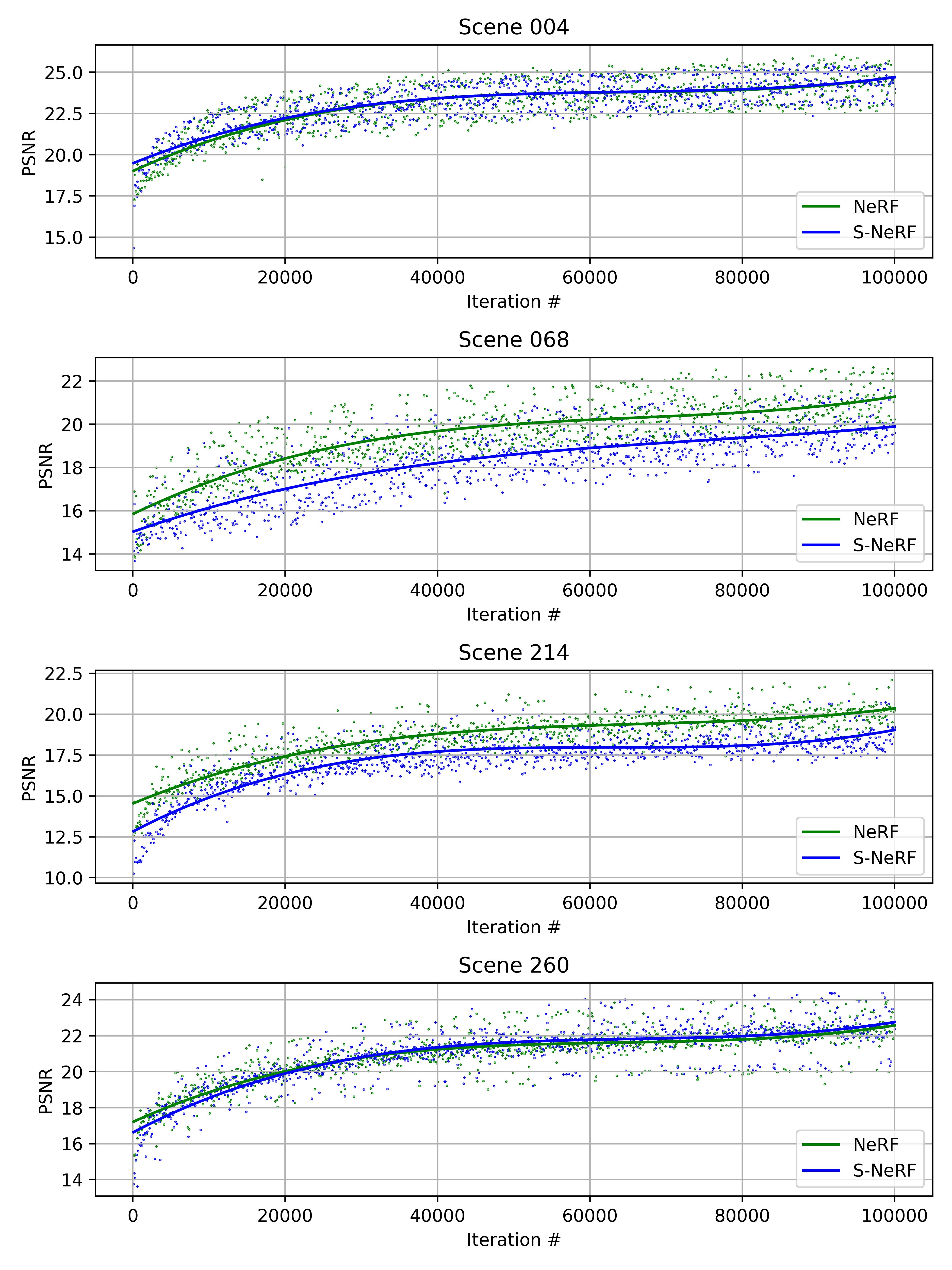}}
  \caption{NeRF and S-NeRF results at 100,000 epochs \label{fig:nerf_snerf}}
\end{figure}

Both the S-NeRF and NeRF models were run on the four scenes until optimal convergence (100k epochs) with viewing angles. The results for the PSNR and SSIM metrics can be observed in Figure \ref{fig:nerf_snerf}, while the qualitative results can be observed in \ref{fig:qualitative_nerf_snerf}. In this case, positional encoders of size 10 for the locations, and 4 for the viewing angles were used for both models. The results from S-NeRF are quantitatively similar in terms of PSNR and SSIM metrics, but they show promise qualitatively, as can be particularly noticed in the improved lighting and shadows in the qualitative results in scene 260 (bottom right in Figure \ref{fig:qualitative_nerf_snerf}).

S-NeRF evaluates the quality of surface extraction on its maximal altitude. The loss is computed using Mean Average Error (MAE) between the expected and ground truth altitude that is measured by the airborne LiDAR. The estimated altitude is represented as a weighted sum of the sampled altitude with its respective weights. The first visible surface object is assigned a weight close to 1. 

The depth maps representing the scenes' altitudes are shown in Figure \ref{fig:depth_maps}. From these, we noticed that NeRF and S-NeRF struggle with flat surfaces, as for the water surface in scene 260 or the mostly flat scene 004. On the other hand, the altitude is correctly reconstructed for taller buildings. 

%-------------------------------------------------------------------------
\section{Conclusion}

We started from an implementation of the NeRF model and developed a data pipeline in order to test it on satellite images. A recent approach based on the S-NeRF model was integrated, directly modeling the lighting conditions together with the radiance. Both models were tuned extensively in order to produce the final results.

We limited our exploration to the NeRF and S-NeRF implementations, but there are clear next steps for improvement. NeRF-W has recently received particular attention for its robustness with occlusions and lighting differences within the training images, which are attractive features for the satellite imagery application. In addition, S-NeRF showed success in using SIREN as its backbone model instead of an MLP with ReLUs, which would be interesting to compare against. Finally, any NeRF-based model struggles with flat surfaces with scarce texture or visual cues. An example in our dataset is represented by the misinterpreted water surface in scene 260. It would be interesting to combine an alternative semantic segmentation approach to bias the loss to reduce this altitude estimation error.

%-------------------------------------------------------------------------
% References
\clearpage
{\small
\bibliographystyle{ieee_fullname}
\bibliography{egbib}

\begin{thebibliography}{10}\itemsep=-1pt

\bibitem{adamkiewicz2021vision}
Michal Adamkiewicz, Timothy Chen, Adam Caccavale, Rachel Gardner, Preston
  Culbertson, Jeannette Bohg, and Mac Schwager.
\newblock Vision-only robot navigation in a neural radiance world.
\newblock {\em arXiv preprint arXiv:2110.00168}, 2021.

\bibitem{andrew2001multiple}
Alex~M Andrew.
\newblock Multiple view geometry in computer vision.
\newblock {\em Kybernetes}, 2001.

\bibitem{boykov2001fast}
Yuri Boykov, Olga Veksler, and Ramin Zabih.
\newblock Fast approximate energy minimization via graph cuts.
\newblock {\em IEEE Transactions on pattern analysis and machine intelligence},
  23(11):1222--1239, 2001.

\bibitem{derksen2021shadow}
Dawa Derksen and Dario Izzo.
\newblock Shadow neural radiance fields for multi-view satellite
  photogrammetry.
\newblock In {\em Proceedings of the IEEE/CVF Conference on Computer Vision and
  Pattern Recognition}, pages 1152--1161, 2021.

\bibitem{c6tm-vw12-19}
Bertrand Le~Saux, Naoto Yokoya, Ronny Hänsch, and Myron Brown.
\newblock Data fusion contest 2019 (dfc2019), 2019.

\bibitem{li20213d}
Yunzhu Li, Shuang Li, Vincent Sitzmann, Pulkit Agrawal, and Antonio Torralba.
\newblock 3d neural scene representations for visuomotor control.
\newblock {\em arXiv preprint arXiv:2107.04004}, 2021.

\bibitem{lin2021barf}
Chen-Hsuan Lin, Wei-Chiu Ma, Antonio Torralba, and Simon Lucey.
\newblock Barf: Bundle-adjusting neural radiance fields.
\newblock {\em arXiv preprint arXiv:2104.06405}, 2021.

\bibitem{martin2021nerf}
Ricardo Martin-Brualla, Noha Radwan, Mehdi~SM Sajjadi, Jonathan~T Barron,
  Alexey Dosovitskiy, and Daniel Duckworth.
\newblock Nerf in the wild: Neural radiance fields for unconstrained photo
  collections.
\newblock In {\em Proceedings of the IEEE/CVF Conference on Computer Vision and
  Pattern Recognition}, pages 7210--7219, 2021.

\bibitem{mildenhall2020nerf}
Ben Mildenhall, Pratul~P Srinivasan, Matthew Tancik, Jonathan~T Barron, Ravi
  Ramamoorthi, and Ren Ng.
\newblock Nerf: Representing scenes as neural radiance fields for view
  synthesis.
\newblock In {\em European conference on computer vision}, pages 405--421.
  Springer, 2020.

\bibitem{rahaman2019spectral}
Nasim Rahaman, Aristide Baratin, Devansh Arpit, Felix Draxler, Min Lin, Fred
  Hamprecht, Yoshua Bengio, and Aaron Courville.
\newblock On the spectral bias of neural networks.
\newblock In {\em International Conference on Machine Learning}, pages
  5301--5310. PMLR, 2019.

\bibitem{sitzmann2020implicit}
Vincent Sitzmann, Julien Martel, Alexander Bergman, David Lindell, and Gordon
  Wetzstein.
\newblock Implicit neural representations with periodic activation functions.
\newblock {\em Advances in Neural Information Processing Systems}, 33, 2020.

\bibitem{triggs1999bundle}
Bill Triggs, Philip~F McLauchlan, Richard~I Hartley, and Andrew~W Fitzgibbon.
\newblock Bundle adjustment—a modern synthesis.
\newblock In {\em International workshop on vision algorithms}, pages 298--372.
  Springer, 1999.

\bibitem{wang2021nerf}
Zirui Wang, Shangzhe Wu, Weidi Xie, Min Chen, and Victor~Adrian Prisacariu.
\newblock Nerf--: Neural radiance fields without known camera parameters.
\newblock {\em arXiv preprint arXiv:2102.07064}, 2021.

\bibitem{lin2020nerfpytorch}
Lin Yen-Chen.
\newblock Nerf-pytorch.
\newblock \url{https://github.com/yenchenlin/nerf-pytorch/}, 2020.

\bibitem{yu2021pixelnerf}
Alex Yu, Vickie Ye, Matthew Tancik, and Angjoo Kanazawa.
\newblock pixelnerf: Neural radiance fields from one or few images.
\newblock In {\em Proceedings of the IEEE/CVF Conference on Computer Vision and
  Pattern Recognition}, pages 4578--4587, 2021.

\end{thebibliography}
}

%-------------------------------------------------------------------------
\appendix
\renewcommand\thefigure{\thesubsection.\arabic{figure}}
\renewcommand{\thetable}{\thesubsection.\arabic{table}}
\renewcommand{\theequation}{\thesubsection.\arabic{equation}}
\section{Work Division}

\begin{table}[!htb]
\caption{Summary of work division \label{tab:contributions}}
\begin{adjustbox}{width=\columnwidth,center}
\begin{tabular}{|c|l|}
\hline
\textbf{Student Name}        & \multicolumn{1}{c|}{\textbf{Yi Zhang}}                                                                                                                                                                                                                                                                                                                                                   \\ \hline
\textbf{Contributions} & \begin{tabular}[c]{@{}l@{}}General project conceptualization, Debugging help, \\ Altitude estimation functions, Report review\end{tabular}                                                                                                                                                                                                                                               \\ \hline
\textbf{Details}             & \begin{tabular}[c]{@{}l@{}}Defined SSIM, altitude metrics, Experimented SNerf \\ calculated altitude vs. traditional NeRF altitude\\ Implemented multiple focals and images batch loading, \\ Attempted to merge Nerf–learn pose and focal model to S-Nerf\end{tabular}                                                                                                                  \\ \hline
\textbf{Student Name}        & \multicolumn{1}{c|}{\textbf{Wenying Wu}}                                                                                                                                                                                                                                                                                                                                                 \\ \hline
\textbf{Contributions} & \begin{tabular}[c]{@{}l@{}}General project conceptualization, Early research, \\ prototyping + experiments, Debugging help, \\ Data augmentation study, Report review\end{tabular}                                                                                                                                                                                                       \\ \hline
\textbf{Details}             & \begin{tabular}[c]{@{}l@{}}Loss function definition for S-Nerf, Data augmentation, \\ Experimented SNerf calculated altitude vs. NeRF altitude \\ Tried out Nerf– on satellite images (results not satisfying)\end{tabular}                                                                                                                                                              \\ \hline
\textbf{Student Name}        & \multicolumn{1}{c|}{\textbf{Patrick Carroll}}                                                                                                                                                                                                                                                                                                                                            \\ \hline
\textbf{Contributions} & \begin{tabular}[c]{@{}l@{}}General project conceptualization, Primary code developer, \\ Report review\end{tabular}                                                                                                                                                                                                                                                                      \\ \hline
\textbf{Details}             & \begin{tabular}[c]{@{}l@{}}Adapted PyTorch version of NeRF to read in camera \\ extrinsics / intrinsics from the SNerf paper dataset. \\ Adapted PyTorch version of NeRF to implement S-Nerf.  \\ Significant prototyping work prior to \\ implementation to define final project proposal and work items. \\ Tried out Nerf– on satellite images (results not satisfying)\end{tabular} \\ \hline
\textbf{Student Name}        & \multicolumn{1}{c|}{\textbf{Federico Semeraro}}                                                                                                                                                                                                                                                                                                                                          \\ \hline
\textbf{Contributions} & \begin{tabular}[c]{@{}l@{}}General conceptualization, Debugging help, Hyperparameter study,  \\ Jobs scheduling and post-processing results, Report writing\end{tabular}                                                                                                                                                                                                                 \\ \hline
\textbf{Details}             & \begin{tabular}[c]{@{}l@{}}Only person with access to the V100s, which translated \\ into a lot of data wrangling and job scheduling. \\ Helped with debugging implementation errors, \\ hyperparameter tuning and general testing, creating and documenting results\end{tabular}                                                                                                                    \\ \hline
\end{tabular}
\end{adjustbox}
\end{table}
% \clearpage
\section{Qualitative results for NeRF and S-NeRF}\label{sec:hyper}

\begin{figure}[!htb]
      \centerline{\includegraphics[width=0.9\linewidth]{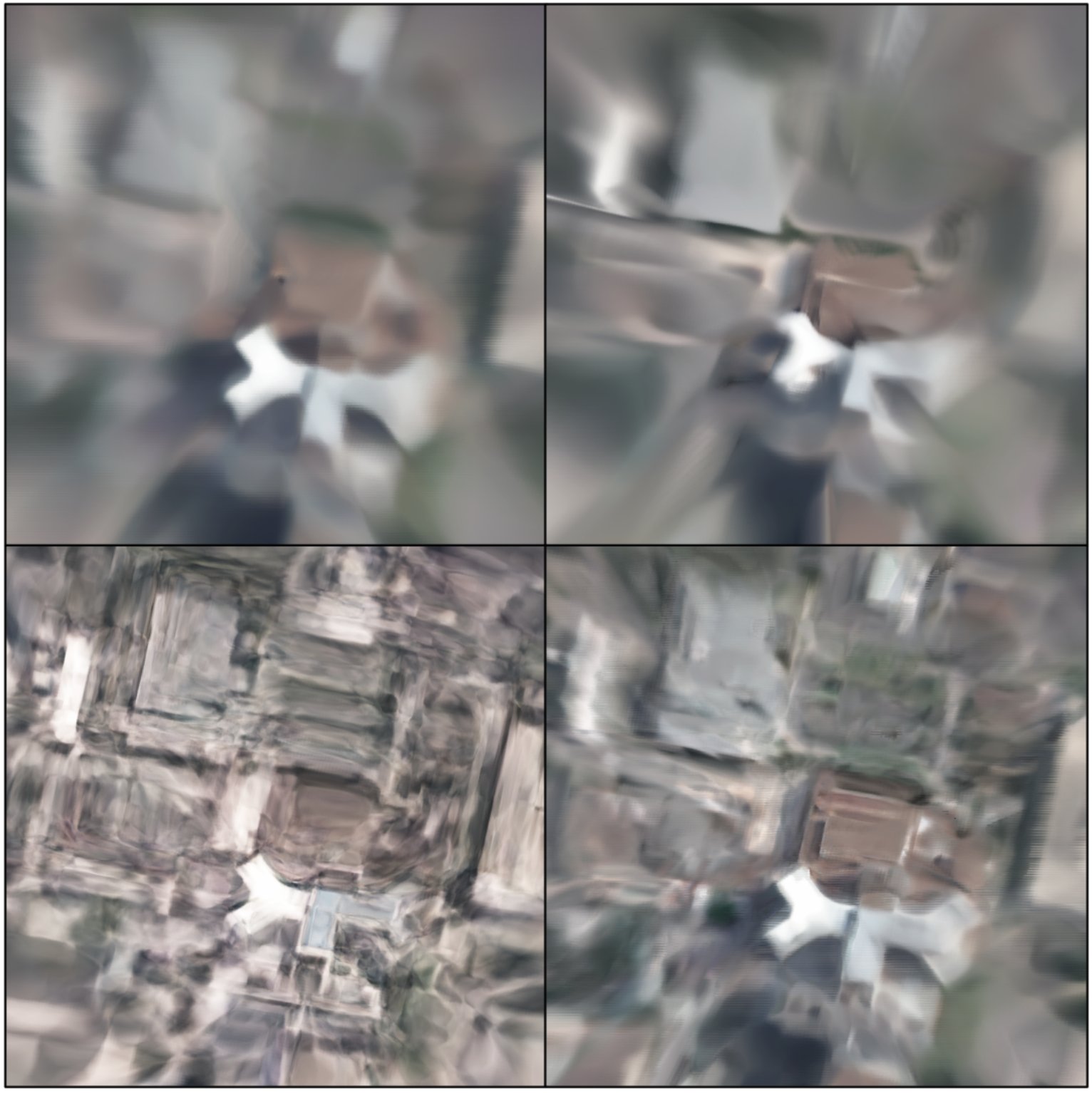}}
  \caption{Qualitative results for scene 068 between using batching of rays at random throughout images (top left) or one image at a time (top right), or using the viewing angle (bottom left) or only using the ray sample coordinates (bottom right). Note that the two images on top correspond to the NeRF model trained until 10,000 epochs, while the bottom two were trained until 100,000 \label{fig:batching_viewdirs}}
\end{figure}

% \begin{landscape}
% \centering
% \vspace*{\fill}
% \begin{figure}[htpb]
%   \centering
%   \includegraphics[width=1.3\textwidth]{figures/qualitative_hyper.jpg}
%   \caption{Novel views for the hyperparameter study for the basic NeRF implementation for the four rendered scenes at 10,000 iterations \label{fig:qualitative_hyper}}
% \end{figure}
% \vfill
% \end{landscape}

\begin{figure*}
  \centerline{\includegraphics[height=\textheight]{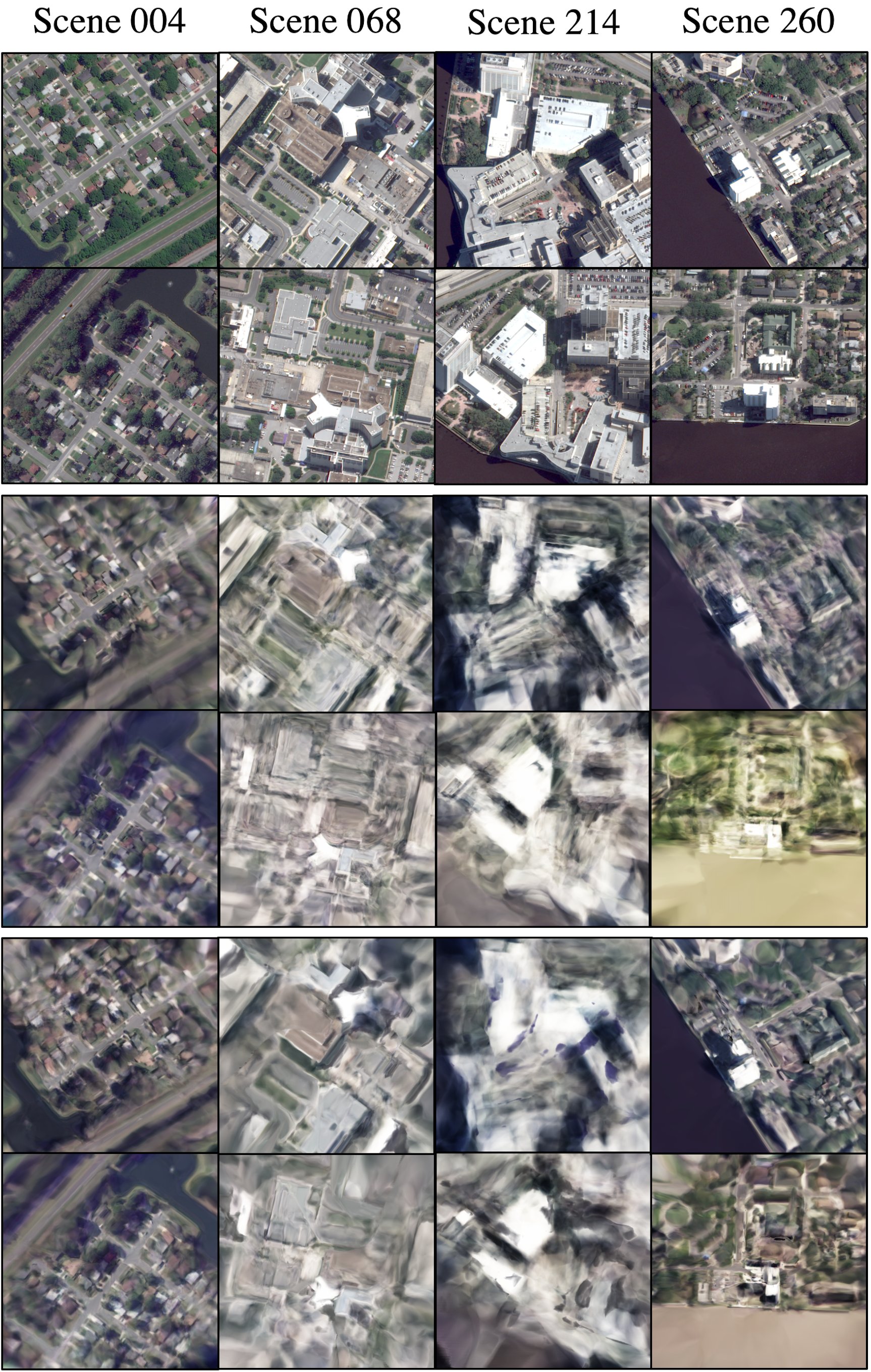}}
  \caption{Ground truth (top), NeRF (middle), and S-NeRF (bottom) views at 100,000 epochs \label{fig:qualitative_nerf_snerf}}
\end{figure*}

\begin{figure*}[htpb]
  \centering
  \includegraphics[width=\textwidth,height=\textheight,keepaspectratio]{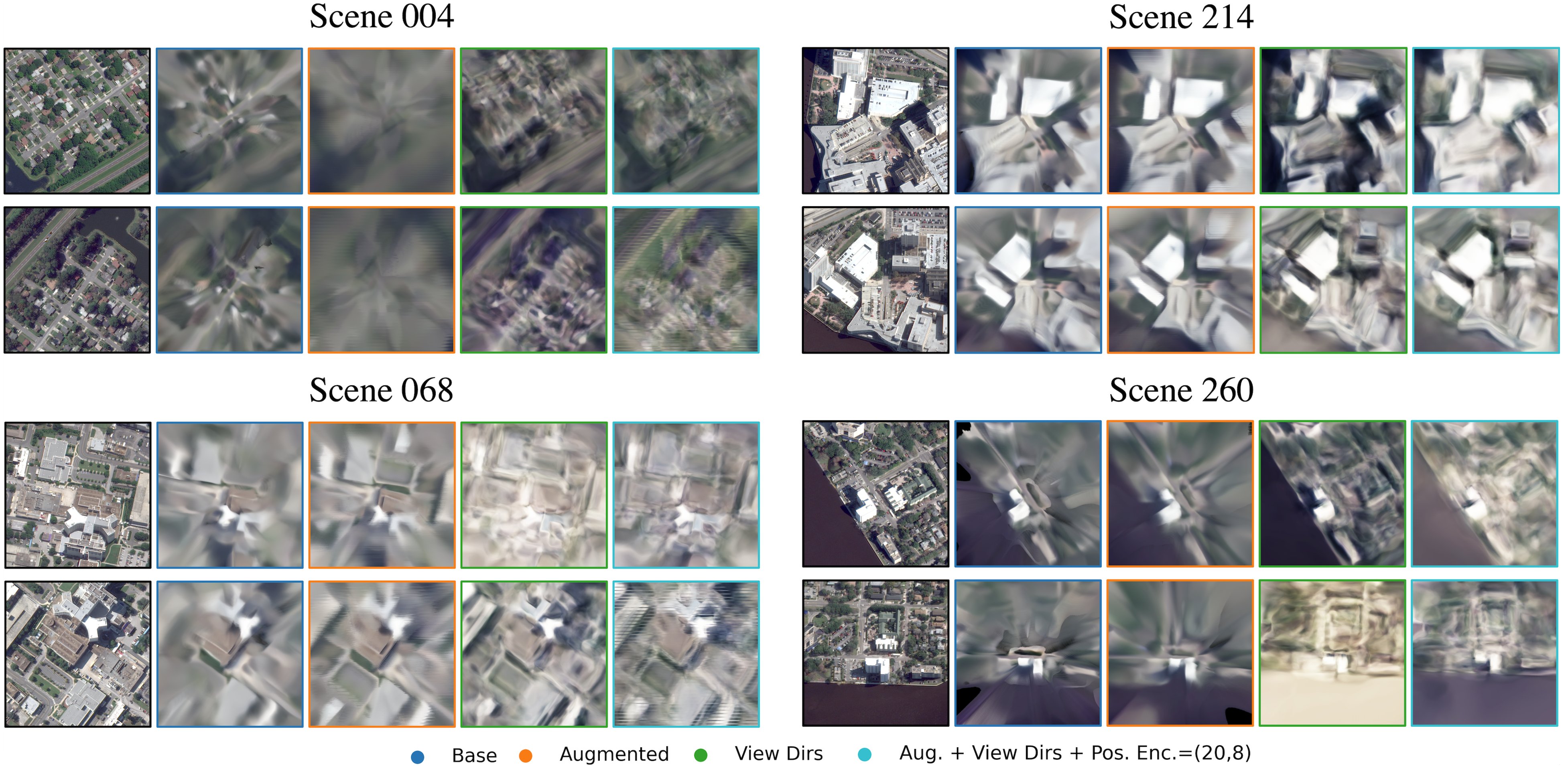}
  \caption{Novel views for the hyperparameter study for the basic NeRF implementation for the four rendered scenes at 10,000 iterations \label{fig:qualitative_hyper}}
\end{figure*}

\begin{figure*}[htpb]
  \centering
  \includegraphics[width=\textwidth,height=\textheight,keepaspectratio]{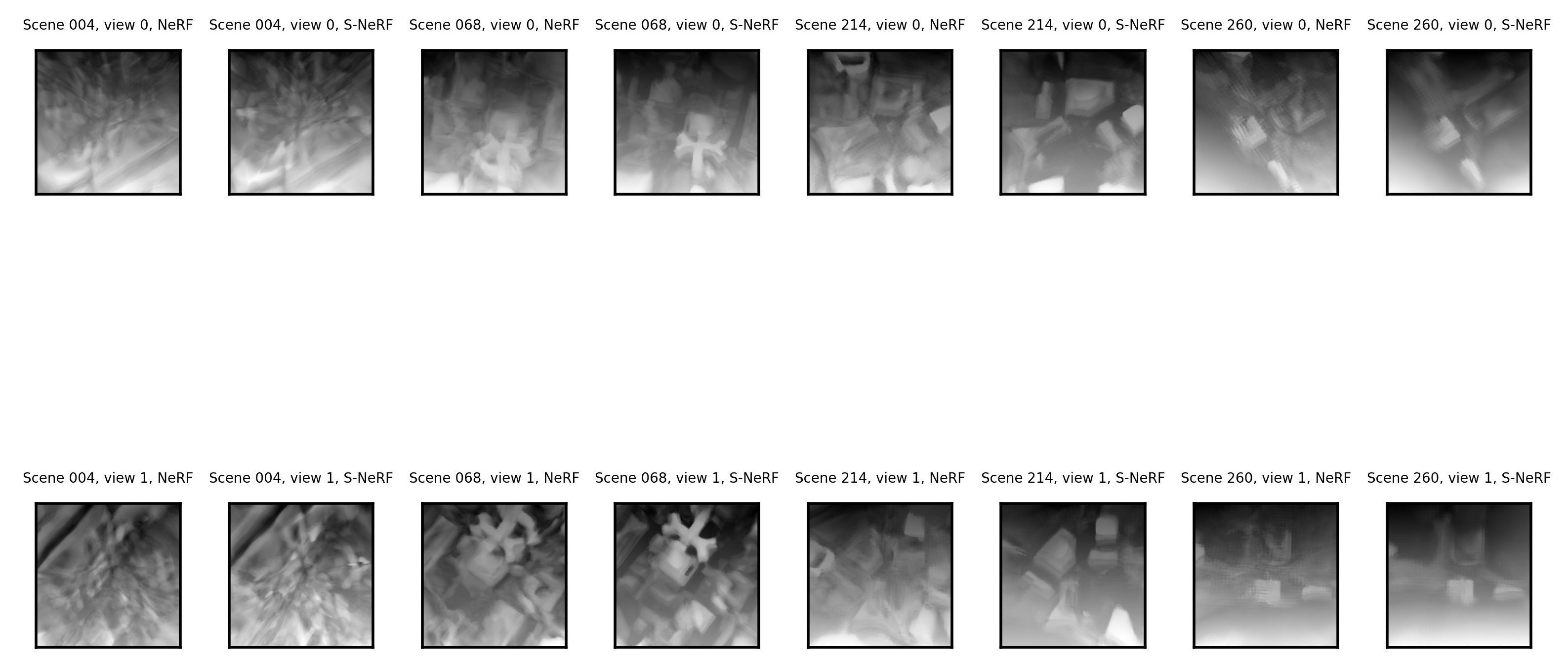}
  \caption{Depth maps at 200,000 epochs \label{fig:depth_maps}}
\end{figure*}

% \begin{landscape}
% \centering
% \vspace*{\fill}
% \begin{figure}[htpb]
%   \centering
%   \includegraphics[width=1.2\textwidth]{figures/depth1.jpg}
%   \caption{Depth maps at 200,000 epochs \label{fig:depth_maps}}
% \end{figure}
% \vfill
% \end{landscape}

\end{document}